\documentclass{article}
\usepackage{spconf,amsmath,graphicx}

\usepackage{multirow}
\usepackage{stfloats}
\usepackage{subfigure} 
\usepackage{sidecap}
\usepackage{float}
\usepackage{bm}

\title{Using Feature Alignment Can Improve Clean Average Precision and Adversarial Robustness in Object Detection}
\name{Weipeng Xu $^{1}$ \thanks{$^*$Corresponding author: Shaoyou Pan, pansy@ssfjd.cn. This study was supported by grants from Central-level Research Institutes Public Welfare Project (GY2017G-6); and from the Science and Technology Committee of Shanghai Municipality (19DZ2292700).} \qquad Hongcheng Huang $^{1}$ \qquad Shaoyou Pan $^{2, *}$}

\address{$^{1}$ Shanghai Jiao Tong University, China \\
		$^{2}$ Shanghai Forensic Service Platform, Academy of Forensic Science, China}
\begin{document}
	%
	\maketitle
	\begin{abstract}
		The 2D object detection in clean images has been a well studied topic, but its vulnerability against adversarial attack is still worrying. Existing work \cite{9009990} has improved robustness of object detectors by adversarial training, at the same time, the average precision (AP) on clean images drops significantly. In this paper, we propose that using feature alignment of intermediate layer can improve clean AP and robustness in object detection. Further, on the basis of adversarial training, we present two feature alignment modules: Knowledge-Distilled Feature Alignment (KDFA) module and Self-Supervised Feature Alignment (SSFA) module, which can guide the network to generate more effective features. We conduct extensive experiments on PASCAL VOC and MS-COCO datasets to verify the effectiveness of our proposed approach. The code of our experiments is available at https://github.com/grispeut/Feature-Alignment.git.
	\end{abstract}
	\section{Introduction}
	\label{sec:intro}
	
	It is a long-standing problem that deep learning models are easily attacked. Studies have shown that both classification models and detection models are easily defeated by adversarial disturbances \cite{DBLP:journals/corr/SzegedyZSBEGF13, DBLP:journals/corr/GoodfellowSS14, 8237415, DBLP:conf/bmvc/LiTCBL18, ijcai2019-134}. In order to improve robustness of deep learning models, the mainstream method is based on adversarial training \cite{DBLP:journals/corr/GoodfellowSS14, DBLP:conf/iclr/MadryMSTV18}. Adversarial training is to solve a two-level optimization problem. The inner loop is a maximization problem that maximizes the proxy loss to generate adversarial samples. The outer loop is a minimization problem that minimizes the loss to train the parameters of models.  
	
	On the basis of adversarial training, some researchers promote adversarial robustness from the perspective of generating adversarial samples, such as reusing gradient information to generate adversarial samples \cite{NEURIPS2019_7503cfac}, FGSM with random initialization \cite{Wong2020Fast} and constructing adversarial samples in an unsupervised manner \cite{NEURIPS2019_d8700cbd}. Some researchers also optimize the minimization problem of outer loop to increase robustness, such as adding a self-supervised branch to the classification network, which uses the output of the penultimate layer as the input of this branch and predicts the rotation angle of the picture \cite{NEURIPS2019_a2b15837}, or adding the regularization term GradAlign to prevent catastrophic overfitting \cite{abs-2007-02617}, or applying self-supervised comparative representation learning to adversarial training \cite{abs-2006-07589}. 
	
	The above-mentioned methods have been verified on classifiers and benchmark datasets, but few works have extended their methods to object detection task. The reason may be that the object detector is more complicated, which leading to higher training costs, and some methods are not suitable for direct migration to the adversarial training of object detection, such as FGSM with random initialization using cyclic learning rate to accelerate the convergence speed \cite{Wong2020Fast}, the self-supervised branch to predict rotation \cite{NEURIPS2019_a2b15837}, and GradAlign which needs to create calculation graph of the gradient of input \cite{abs-2007-02617}, resulting in the batch size should be reduced under limited graphics memory. Nevertheless, the discovery of catastrophic overfitting phenomenon \cite{Wong2020Fast, abs-2007-02617} and the combining unsupervised or self-supervised learning with adversarial training \cite{NEURIPS2019_d8700cbd, abs-2006-07589} are still instructive for object detection task.

	For the robustness of detector, Zhang et al. \cite{9009990} propose to take object detection as multi-task learning and use task oriented domain to generate adversarial samples. They conduct experiments on a variety of one-stage detectors and the experiments show that their method can improve robustness of models. However, we find that the clean AP drops significantly (about 25 points on PASCAL VOC), while the robustness against attack is improved. In our experiments, we find that the clean AP and robustness are really difficult to obtain at the same time. Therefore, we hope to improve robustness on the basis of reducing the drop of clean AP.
	
	In this paper, we are devoted to better balance clean AP and robustness. Considering that detectors have more layers than classifiers, there are additional location branches and feature fusion modules such as feature pyramid modules \cite{8099589}. Therefore, in the face of a deeper and more complex detection network, we propose  feature alignment approach, which guide the output of intermediate feature layer to equal to prior features. Then we provide two guiding methods: Knowledge-Distilled Feature Alignment (KDFA) and Self-Supervised Feature Alignment (SSFA). Knowledge distillation is widely used in model compression \cite{DBLP:journals/corr/HintonVD15, DBLP:journals/corr/RomeroBKCGB14, 8953432}, which can transfer the knowledge learned by the teacher network to the student network. Inspired by this, we propose to use the detector trained on clean dataset as the teacher, and distill feature knowledge of middle layer to student network. Then motivated by the work of self-supervised learning \cite{9157636, abs-2002-05709, abs-2006-07589, abs-2011-10566}, we propose SSFA which takes the clean image and its corresponding adversarial sample as different representations of the same data, and maximizes their feature similarity in the middle layer. Compared with self-supervised representation learning, prior features of feature alignment not only come from siamese networks, but also from teacher networks. So feature alignment is a more general concept.

	The contribution of this paper includes two aspects: $\bm{i})$ we propose to guide the output of middle layer to strengthen adversarial training in object detection, and provide two feature alignment methods based on knowledge distillation and self-supervised learning; $\bm{ii})$ on PASCAL VOC and MS-COCO datasets, our approach obtains higher clean AP and robustness than existing works across one-stage detector YOLO-V3 \cite{abs-1804-02767} and two-stage detector FASTER-RCNN-FPN \cite{8099589}.

\section{our approach}

\label{sec:format}

In this section, we first revisit the basic form of adversarial training. Then we analyze the advantage of feature alignment in intermediate layer and introduce our KDFA and SSFA module. Finally, we introduce evaluation methods of clean AP and robustness.

\subsection{Adversarial Training}

Adversarial training is proposed by Goodfellow et al.\cite{DBLP:journals/corr/GoodfellowSS14}, and can be formulated as follows:

\begin{equation}
\begin{split}
\label{minmax_new}
\min _{\theta} \sum_{i} ( (1 - \alpha) * \max _{\delta \in \Delta} \ell\left(f_{\theta}\left(x_{i}+\delta\right), y_{i}\right) \\ +  \alpha * \ell\left(f_{\theta}\left(x_{i}\right), y_{i}\right))
\end{split}
\end{equation}

where $f$, $\theta$, $\ell$ denote the deep learning model, parameters of model and proxy loss function, and $x_{i}$,$y_{i}$,$\delta$,$\Delta$ are the clean sample, the label of sample, disturbances added to clean sample and the upper bound of disturbances. 

\begin{table*}[tbp]
	\centering
	\begin{tabular}{c|c|c|c|c|c|c|c|c|c|c|c}
		\hline \multicolumn{3}{c|} {\multirow{2}{*}  {model performance} }& \multicolumn{3}{c|} {trained with PGD-1} & \multicolumn{3}{c|} {trained with PGD-2} & \multicolumn{3}{c} {trained with PGD-3}\\
		\cline{4-12}
		\multicolumn{3}{c|}{ } & clean AP & advAP & acAP & clean AP & advAP & acAP & clean AP & advAP & acAP \\
		\cline{4-12}
		
		\hline \multirow{5}{*} { FPN } & \multicolumn{2}{c|} { AT \cite{DBLP:journals/corr/GoodfellowSS14} } & 0.834 & 0.074 & 0.454 & 0.824 & 0.236 & 0.530 & 0.822 & 0.137 & 0.480  \\
		\cline{2-3}
		& \multicolumn{2}{c|} {TOD \cite{9009990} }  & 0.833 & 0.106 & 0.470 & 0.828 & 0.247 & 0.538 & 0.821 & 0.173 & 0.497\\
		\cline{2-3}
		& \multirow{3}{*} { ours }& KDFA  & 0.833 & 0.133 & 0.483 & 0.828 & 0.273 & 0.551 & 0.823 & 0.217 & 0.520 \\
		& & SSFA  & 0.833 & 0.123 & 0.478 & 0.829 & 0.267 & 0.548 & 0.824 & 0.190 & 0.507\\
		& & FA  & 0.834 & 0.166 & \textbf{0.500} & 0.832 & 0.286 & \textbf{0.559} & 0.825 & 0.260 & \textbf{0.543}\\
		
		\hline \multirow{5}{*} { YOLO } & \multicolumn{2}{c|} { AT \cite{DBLP:journals/corr/GoodfellowSS14} } & 0.724 & 0.201 & 0.463 & 0.665 & 0.217 & 0.441 & 0.604 & 0.201 & 0.403 \\
		\cline{2-3}
		& \multicolumn{2}{c|} {TOD \cite{9009990} } &   0.702 & 0.208 & 0.455 & 0.659 & 0.228 & 0.444 & 0.581 & 0.206 & 0.394 \\
		\cline{2-3}
		& \multirow{3}{*} { ours }& KDFA & 0.739 & 0.217 & 0.478 & 0.702 & 0.240 & 0.471 & 0.672 & 0.240 & 0.456 \\
		& & SSFA & 0.730 & 0.189 & 0.460 & 0.693 & 0.221 & 0.457 & 0.653 & 0.223 & 0.438\\
		& & FA & 0.745 & 0.217 & \textbf{0.481} & 0.712 & 0.241 & \textbf{0.477} & 0.677 & 0.241 & \textbf{0.459}\\
		
		\hline
		
	\end{tabular}
	\caption{Evaluation results on PASCAL VOC. FPN is FASTER-RCNN-FPN. YOLO is YOLO-V3. AT is vanilla adversarial training. TOD is using task oriented domain in adversarial training. FA is using SSFA + KDFA.}
	\label{res}
\end{table*}

\subsection{Feature Alignment of Middle Layer}

\begin{figure}[tb]
	
	\centering
	\centerline{\includegraphics[width=8cm]{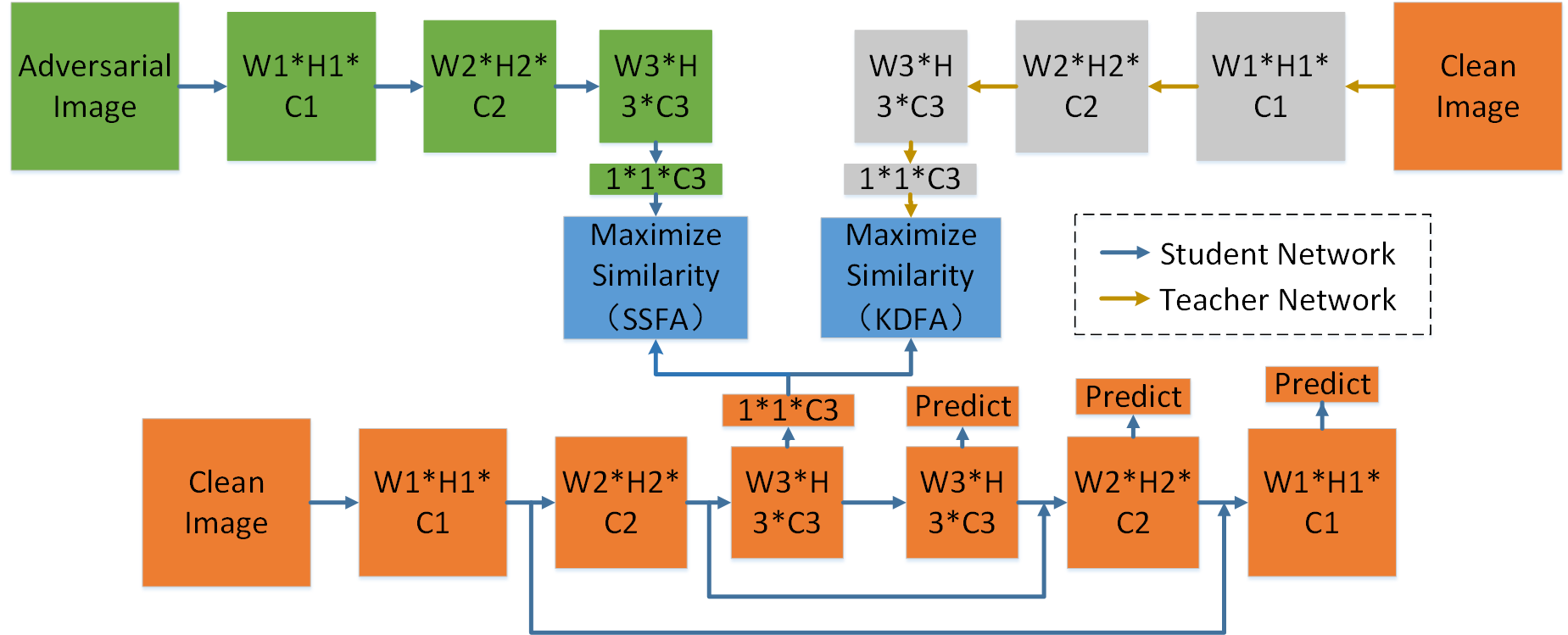}}
	%
	\caption{The structure of feature alignment, consisting of KDFA and SSFA module.}
	\label{struct}
\end{figure}
Reviewing the Spectral Analysis of Unstability proposed by Szegedy et al.\cite{DBLP:journals/corr/SzegedyZSBEGF13}, the unstability of network can be expressed by the Lipschitz constant of each layer, which can be formulated as follows:

\begin{equation}
\begin{split}
\|f(x)-f(x+\delta)\| \leq \prod_{k=1}^{K} L_{k}\|\delta\| \\
\|f_{mid}(x)-f_{mid}(x+\delta)\| \leq \prod_{k=1}^{M} L_{k}\|\delta\|
\end{split}
\end{equation}

where $L_{k}$ is the the upper Lipschitz constant of layer $k$, and $f_{mid}(x)$ denotes the output of middle layer. In this paper, the last layer of backbone is set as the $M$th layer.

By imposing constraints, $f_{mid}(x+\delta)$ can be approximately equal to $f_{mid}(x)$ or $ft_{mid}(x)$, where $ft$ denotes the teacher network.  We define the process of guiding $f_{mid}(x+\delta)$ equal to another prior features as feature alignment. Feature alignment can reduce the unstability of network before $M$th layer, thereby reducing the probability of adversarial examples. In addition, the alignment can guide the learning of the network before $M$th layer to generate more effective features, and effective features are conducive to the learning of the network after $M$th layer. 

The structure of feature alignment is shown in Fig.\ref{struct}. \textbf{KDFA} is the case that guiding $f_{mid}(x+\delta)$ equals to $ft_{mid}(x)$.  We use the detector trained on the clean dataset as the teacher network, of which feature representation outputted by the middle layer on the clean data can be a good prior knowledge. \textbf{SSFA} is the case that guiding $f_{mid}(x+\delta)$ equals to $f_{mid}(x)$, which uses the prior knowledge that clean pictures and their corresponding adversarial examples should have the same feature representation. In addition, considering for each sample, the dimension of output feature is W*H*C, which is a high-dimensional feature representation, and the feature may be noisy or redundant. In our work, we project such a high-dimensional feature into a low-dimensional manifold. Specifically, we use average-pooling as the feature projection operator, so the low-dimensional manifold is a C-dimensional feature. Then feature alignment is achieved by maximizing their feature similarity in the middle layer. Inspired by the idea of stop-grad in SimSiam \cite{abs-2011-10566}, when calculating the feature representation of the clean picture in student network, we do not keep the calculation graph. 

Adversarial training with feature alignment can be formulated as follows:

\begin{equation}
\begin{split}
\label{our_at}
\min _{\theta} \sum_{i}   (1 - \alpha) * \max _{\delta \in \Delta} \ell\left(f_{\theta}\left(x_{i}+\delta\right), y_{i}\right) +  \alpha * \ell\left(f_{\theta}\left(x_{i}\right), y_{i}\right) \\
+ \beta * (1- \operatorname{cos\_sim}(f_{mid}(x_{i}+\delta),f_{mid}(x_{i}))) \\
+ \gamma * (1- \operatorname{cos\_sim}(f_{mid}(x_{i}+\delta),ft_{mid}(x_{i}))) 
\end{split}
\end{equation}

where $\operatorname{cos\_sim}$ denotes cos similarity, which can be expressed as $\operatorname{cos\_sim}(x, y)=x^{T} y /\|x\|\|y\|$.

\subsection{PGD attack}
In order to verify our approach, we use the powerful attack PGD \cite{DBLP:conf/iclr/MadryMSTV18} to test the robustness of detector. The k-step PGD (PGD-k) can be formulated as follows:

\begin{equation}
\begin{split}
x^{t+1}=\operatorname{Proj}_{x+\mathcal{S}}\left(x^{t}+\frac{\varepsilon}{k} \operatorname{sgn}\left(\nabla_{x^{t}} \ell \left(f_{\theta}\left(x^{t}\right), y\right)\right)\right)
\end{split}
\end{equation}

where $\varepsilon$ denotes the total adversarial budget, $k$ and $\frac{\varepsilon}{k}$ are the PGD steps and step size. 

In this paper, $\varepsilon$ is set to $0.03$ when testing adversarial robustness of detectors. The AP we use is COCO-style AP at the IoU threshold of 0.5. Given that the different performance under different PGD-k attacks, We use the average adversarial AP under different steps (1, 3, 5, 10) as the adversarial precision, and denote it as \textbf{advAP}. Then, the mean of advAP and clean AP is used to evaluate the performance of detectors, and we denote it as \textbf{acAP}.

\section{experiments}
\label{sec:exp}

\subsection{Experiment and Implementation Details}

We evaluate our approach on PASCAL VOC and MS-COCO. The PASCAL VOC training set uses the combination of 2007 and 2012 $trainval$, and the test set uses the 2007 $test$.  The MS-COCO training set uses $train+valminusminival$ 2014, and the test set uses $minival$ 2014. 

The detectors we use are the one-stage detector (YOLO-V3) and two-stage detector (FASTER-RCNN-FPN with ResNet-50 as backbone). The $M$th layer in YOLO-V3 is the last layer of Darknet-53. For FASTER-RCNN-FPN it is the last layer of Resnet-50. In adversarial training, the optimizer and hyper-parameter settings are consistent with the normal training. 


\begin{figure}[tbp]
	
	\centering
	\centerline{\includegraphics[width=8.5cm]{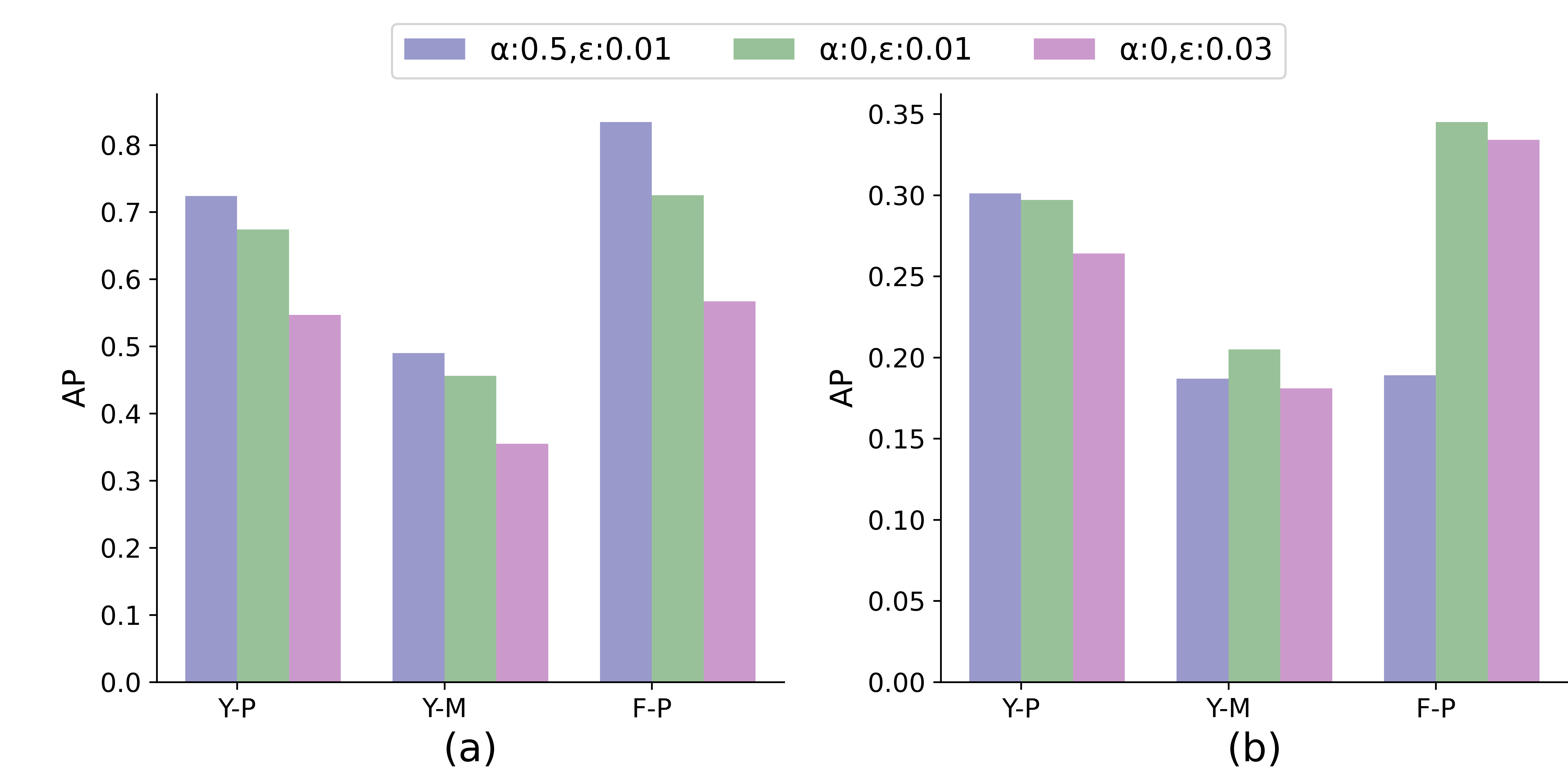}}
	%
	\caption{Model performance trained with PGD-1 under different $\alpha$ and $\varepsilon$. (a) is the performance on clean dataset, and (b) is the performance under PGD-1 attack with 0.03 budget. Y-P is YOLO-V3 on PASCL VOC and Y-M is YOLO-V3 on MS-COCO. F-P is FASTER-RCNN-FPN on PASCL VOC.}
	\label{fig:res}
\end{figure}

\subsection{Results on PASCAL VOC and MS-COCO}

\textbf{Determine the basic form of adversarial training}. We conduct experiments with different $\alpha$ and $\varepsilon$ in vanilla adversarial training (without feature alignment). The results trained with PGD-1 are shown in Fig.\ref{fig:res}. It can be observed that smaller $\varepsilon$ can achieve better result on clean AP and advAP when $\alpha=0$. It may be due to the catastrophic overfitting phenomenon, when large $\varepsilon$ is used in PGD-1 training. We think this is the reason that clean AP drops significantly after adversarial training in the work of Zhang et al. \cite{9009990}. And $\alpha=0.5$ can achieve higher clean AP than $\alpha=0$. Therefore, we set $\varepsilon=0.01, \alpha=0.5$ in next experiments. 

\begin{figure*}[tbp]
	\centering
	\subfigure{
		\begin{minipage}[t]{0.115\linewidth}
			\centering
			\includegraphics[width=2.1cm]{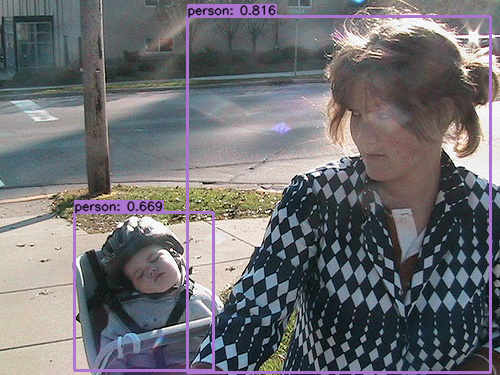}
			\includegraphics[width=2.1cm]{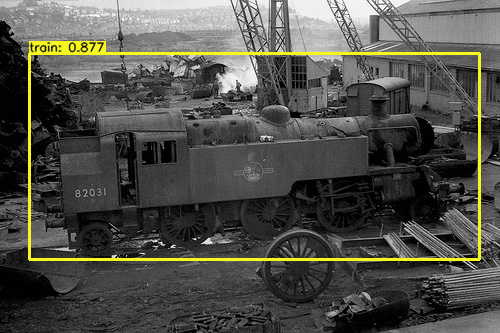}
			\includegraphics[width=2.1cm]{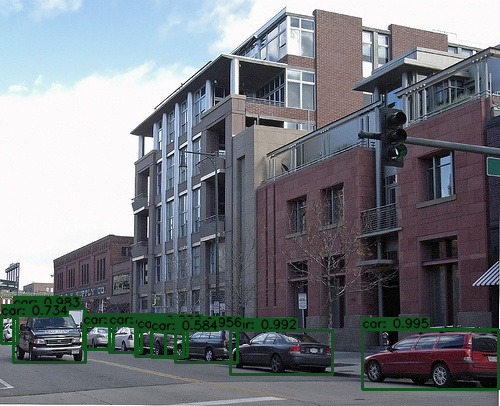}
			\centerline{(a1)}\medskip
		\end{minipage}
		\begin{minipage}[t]{0.115\linewidth}
			\centering
			\includegraphics[width=2.1cm]{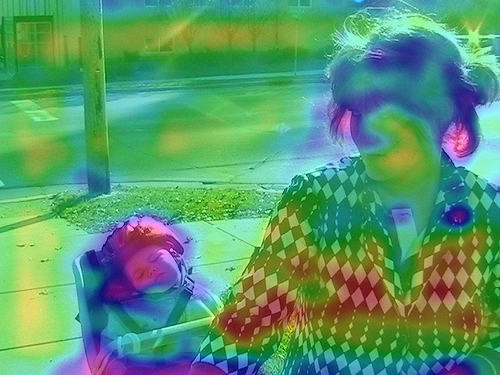}
			\includegraphics[width=2.1cm]{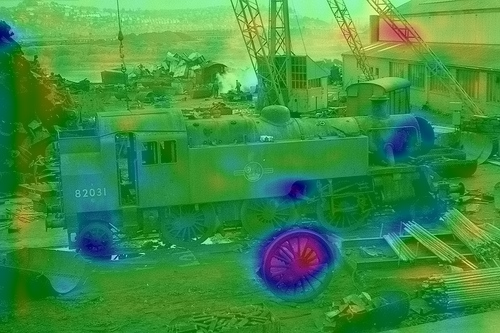}
			\includegraphics[width=2.1cm]{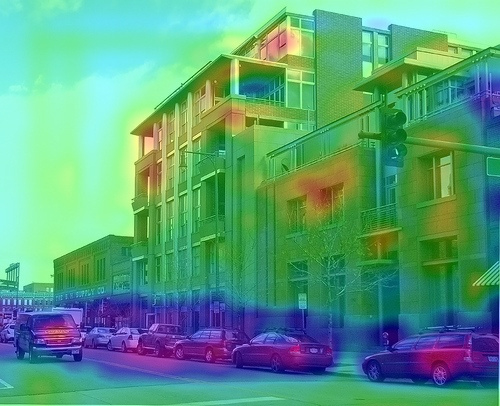}
			\centerline{(a2)}\medskip
		\end{minipage}
		\begin{minipage}[t]{0.115\linewidth}
			\centering
			\includegraphics[width=2.1cm]{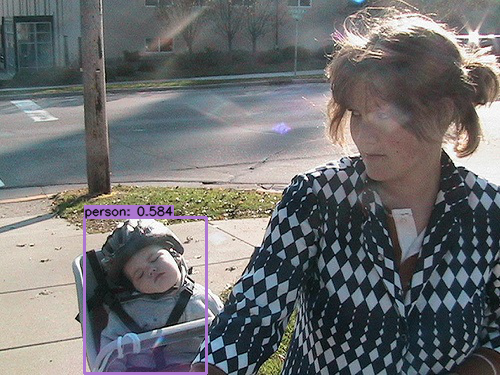}
			\includegraphics[width=2.1cm]{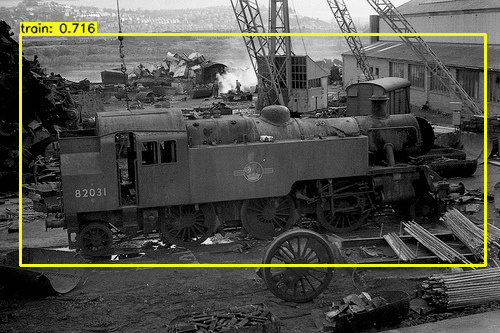}
			\includegraphics[width=2.1cm]{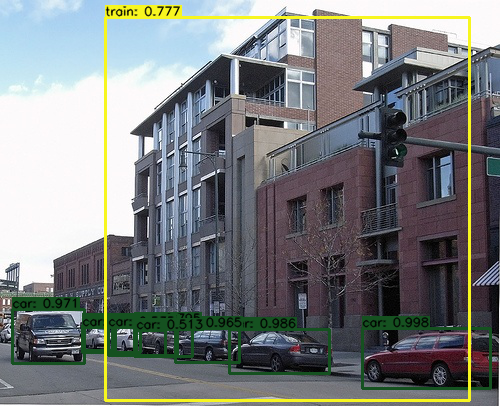}
			\centerline{(b1)}\medskip
		\end{minipage}
		\begin{minipage}[t]{0.115\linewidth}
			\centering
			\includegraphics[width=2.1cm]{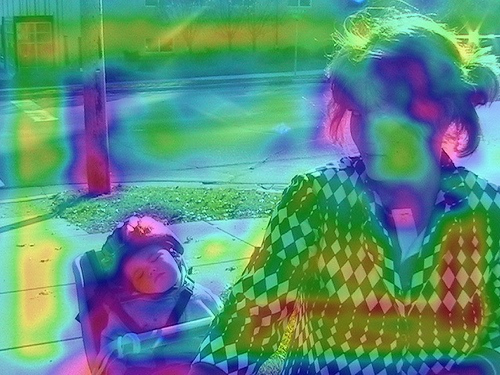}
			\includegraphics[width=2.1cm]{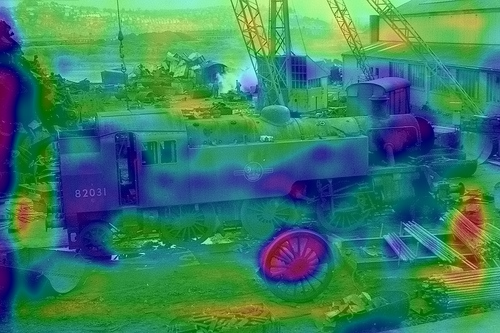}
			\includegraphics[width=2.1cm]{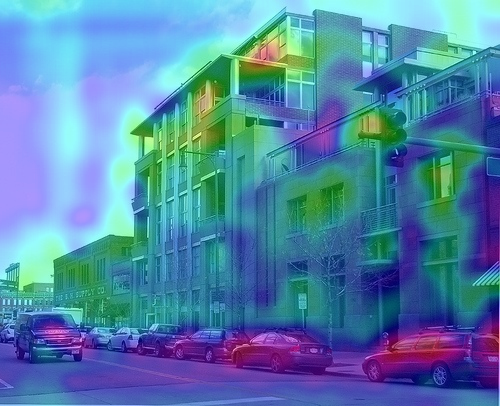}
			\centerline{(b2)}\medskip
		\end{minipage}
		\begin{minipage}[t]{0.115\linewidth}
			\centering
			\includegraphics[width=2.1cm]{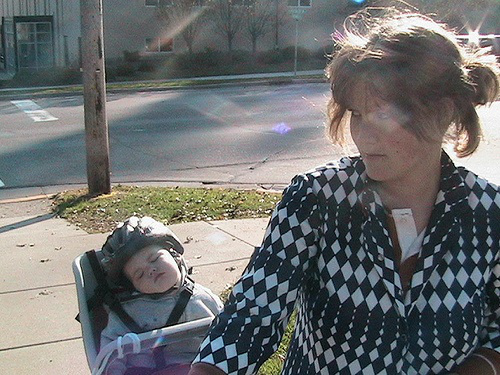}
			\includegraphics[width=2.1cm]{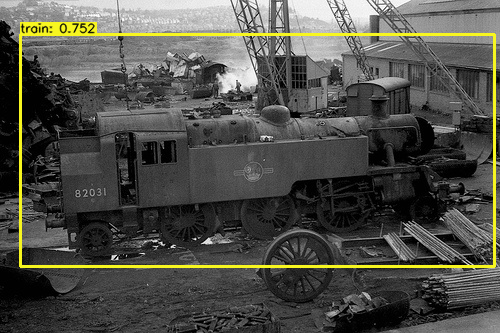}
			\includegraphics[width=2.1cm]{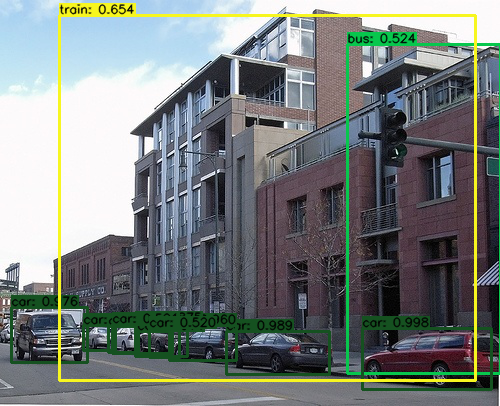}
			\centerline{(c1)}\medskip
		\end{minipage}
		\begin{minipage}[t]{0.115\linewidth}
			\centering
			\includegraphics[width=2.1cm]{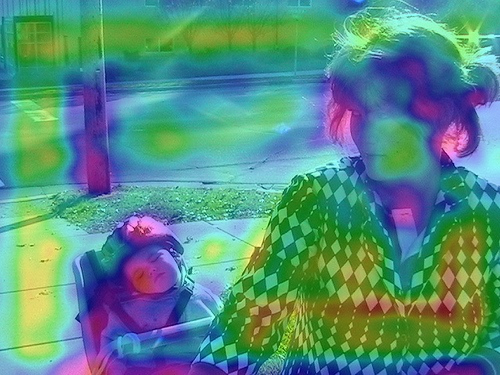}
			\includegraphics[width=2.1cm]{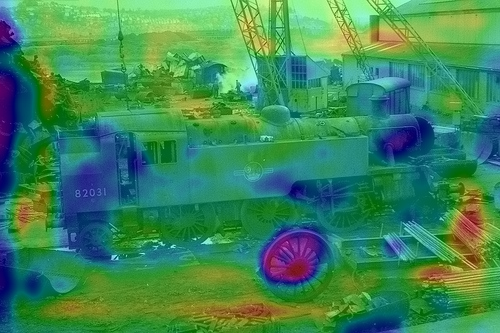}
			\includegraphics[width=2.1cm]{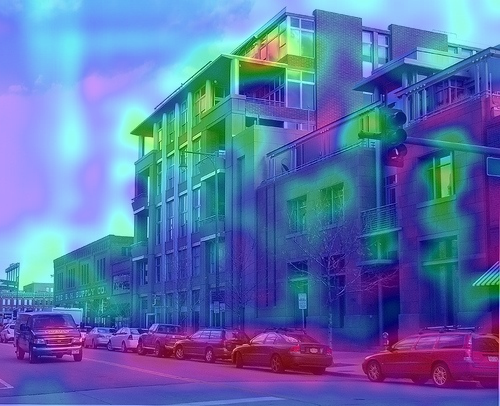}
			\centerline{(c2)}\medskip
		\end{minipage}
		\begin{minipage}[t]{0.115\linewidth}
			\centering
			\includegraphics[width=2.1cm]{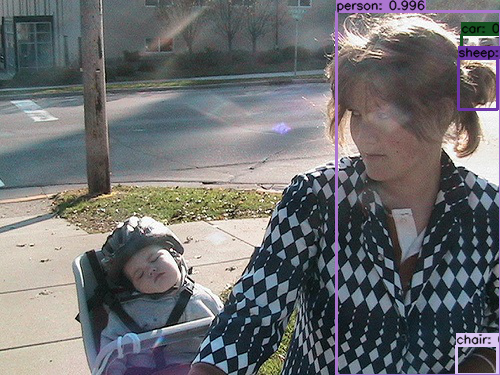}
			\includegraphics[width=2.1cm]{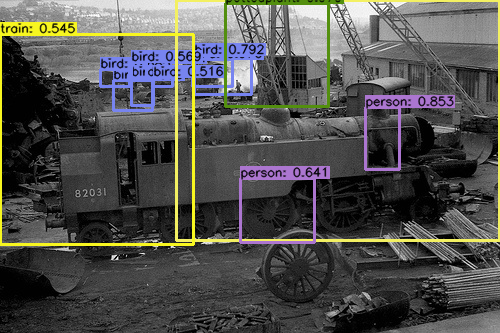}
			\includegraphics[width=2.1cm]{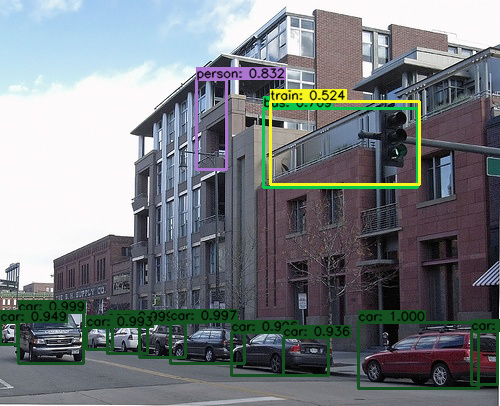}
			\centerline{(d1)}\medskip
		\end{minipage}
		\begin{minipage}[t]{0.115\linewidth}
			\centering
			\includegraphics[width=2.1cm]{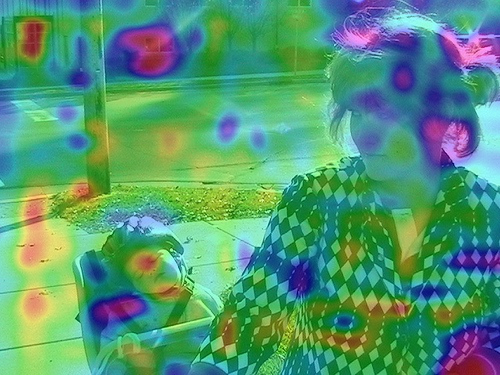}
			\includegraphics[width=2.1cm]{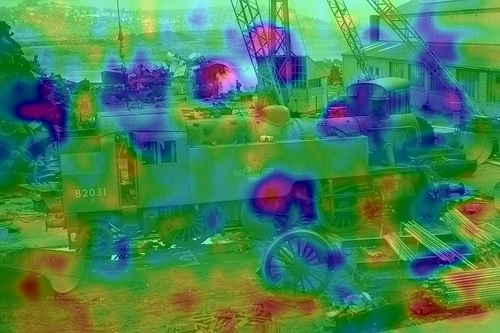}
			\includegraphics[width=2.1cm]{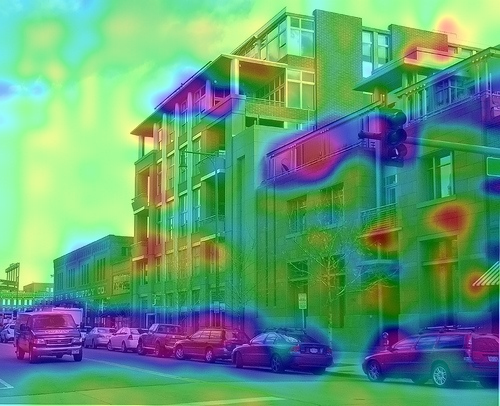}
			\centerline{(d2)}\medskip
		\end{minipage}	
	}
	\centering
	\caption{Detection results and feature visualizations. (a1),(b1),(c1),(d1) are detection results, and (a2),(b2),(c2),(d2) are feature visualizations. (a1) and (a2) are results of SSFA+KDFA. (b1) and (b2) are results of vanilla adversarial training. (c1) and (c2) are results of using task oriented domain. (d1) and (d2) are results of standard settings. Zoom in for more visual details.}
	\label{fv}
\end{figure*}

\begin{table}[tbp]
	\centering
	
	\begin{tabular}{c|c|c|c|c}
		\multicolumn{2}{c|} {model performance} & clean AP & advAP & acAP \\
		\hline
		\multicolumn{2}{c|} {standard}  & \textbf{0.545} & 0.065 & 0.305 \\
		\multicolumn{2}{c|} {AT \cite{DBLP:journals/corr/GoodfellowSS14}} & 0.49 & 0.089 & 0.290 \\
		\multicolumn{2}{c|} {TOD \cite{9009990}} & 0.488 & 0.088 & 0.288 \\
		\cline{1-2}
		\multirow{3}{*} { ours }& KDFA & 0.506 & 0.110 & 0.308 \\
		& SSFA & 0.499 & 0.103 & 0.301 \\
		& FA & 0.510 & \textbf{0.120} & \textbf{0.315} \\
		
	\end{tabular}
	\caption{YOLO-V3 trained with PGD-1 on MS-COCO}
	\label{coco result}
\end{table}

When conducting experiments with feature alignment, $\beta=10, \gamma=10$ are set for YOLO-V3, $\beta=1, \gamma=1$ for FASTER-RCNN-FPN. We update the gradient of network parameters at the first iteration of generating adversarial samples, and keep the output of middle feature layer which is used for SSFA. In addition, in order to quickly improve the robustness on the basis of reducing the drop in clean AP, we first train a based model on the clean dataset for adversarial training, which can accelerate convergence of training. This based model also serves as the teacher network in KDFA. 

\textbf{On PASCAL VOC}, we use different PGD steps (1,2,3) with a fixed step size of 0.01 in adversarial training. The experimental results are shown in Table \ref{res}. It is observed that feature alignment can improve acAP over existing works significantly, and consistently for different setting of adversarial training. In addition, the proposed approach demonstrates the applicability across one-stage and two-stage detector.  Although the clean AP of using task oriented domain has improved largely than results in the work of Zhang et al. \cite{9009990} when $\varepsilon=0.01$, $\alpha=0.5$, our approach still obtains higher clean AP and robustness than using task oriented domain. When using PGD-3 in adversarial training, YOLO-V3 and FASTER-RCNN-FPN both achieve a higher acAP score at least 5 points than vanilla adversarial training, and a higher acAP score at least 4 points than using task oriented domain.

\textbf{On MS-COCO}, the result of YOLO-V3 trained with PGD-1 under PGD attack is shown as Table \ref{coco result}. It further demonstrates the effectiveness of our approach on improving clean AP and robustness.

\subsection{Feature visualization}

To better understand the effect of feature alignment, we visualize the feature of $M$th layer of FASTER-RCNN-FPN with Grad-CAM \cite{DBLP:journals/corr/SelvarajuDVCPB16}. Visualization of test examples under PGD-1 attack ($\varepsilon=0.03$) is shown as Fig.\ref{fv}. It is observed that adversarial training can help to reduce FP and FN than standard training. Our approach can further reduce FN, FP and increase confidence of TP than vanilla adversarial training and using task oriented domain. In feature visualization results, we find that feature alignment can generate more effective features, which focus more on objects.

\section{conclusions}
\label{sec:ref}

In this paper, we present a new approach feature alignment to strengthen adversarial training by guiding output of intermediate feature layer, and propose two feature alignment modules based on knowledge distillation and self-supervised learning, which can help to generate more effective features. On PASCAL VOC and MS-COCO, our approach obtains better performance than existing works. However, there is still a large performance gap between adversarial samples and clean samples. The future work will include experiments on different $M$th layers and more detectors, and explore the balance between different feature alignment modules and adversarial training.

\vfill
\pagebreak


\begin{thebibliography}{10}

\bibitem{9009990}
H.~{Zhang} and J.~{Wang},
\newblock ``Towards adversarially robust object detection,''
\newblock in {\em 2019 IEEE/CVF International Conference on Computer Vision
  (ICCV)}, 2019.

\bibitem{DBLP:journals/corr/SzegedyZSBEGF13}
Christian Szegedy, Wojciech Zaremba, Ilya Sutskever, Joan Bruna, Dumitru Erhan,
  Ian~J. Goodfellow, and Rob Fergus,
\newblock ``Intriguing properties of neural networks,''
\newblock in {\em International Conference on Learning Representations}, 2014.

\bibitem{DBLP:journals/corr/GoodfellowSS14}
Ian~J. Goodfellow, Jonathon Shlens, and Christian Szegedy,
\newblock ``Explaining and harnessing adversarial examples,''
\newblock in {\em International Conference on Learning Representations}, 2015.

\bibitem{8237415}
C.~{Xie}, J.~{Wang}, Z.~{Zhang}, Y.~{Zhou}, L.~{Xie}, and A.~{Yuille},
\newblock ``Adversarial examples for semantic segmentation and object
  detection,''
\newblock in {\em 2017 IEEE International Conference on Computer Vision
  (ICCV)}, 2017.

\bibitem{DBLP:conf/bmvc/LiTCBL18}
Yuezun Li, Daniel Tian, Ming{-}Ching Chang, Xiao Bian, and Siwei Lyu,
\newblock ``Robust adversarial perturbation on deep proposal-based models,''
\newblock in {\em British Machine Vision Conference 2018}. 2018, {BMVA} Press.

\bibitem{ijcai2019-134}
Xingxing Wei, Siyuan Liang, Ning Chen, and Xiaochun Cao,
\newblock ``Transferable adversarial attacks for image and video object
  detection,''
\newblock in {\em Proceedings of the Twenty-Eighth International Joint
  Conference on Artificial Intelligence, {IJCAI-19}}. 7 2019, International
  Joint Conferences on Artificial Intelligence Organization.

\bibitem{DBLP:conf/iclr/MadryMSTV18}
Aleksander Madry, Aleksandar Makelov, Ludwig Schmidt, Dimitris Tsipras, and
  Adrian Vladu,
\newblock ``Towards deep learning models resistant to adversarial attacks,''
\newblock in {\em International Conference on Learning Representations}. 2018,
  OpenReview.net.

\bibitem{NEURIPS2019_7503cfac}
Ali Shafahi, Mahyar Najibi, Mohammad~Amin Ghiasi, Zheng Xu, John Dickerson,
  Christoph Studer, Larry~S Davis, Gavin Taylor, and Tom Goldstein,
\newblock ``Adversarial training for free!,''
\newblock in {\em Advances in Neural Information Processing Systems}, 2019.

\bibitem{Wong2020Fast}
Eric Wong, Leslie Rice, and J.~Zico Kolter,
\newblock ``Fast is better than free: Revisiting adversarial training,''
\newblock in {\em International Conference on Learning Representations}, 2020.

\bibitem{NEURIPS2019_d8700cbd}
Haichao Zhang and Jianyu Wang,
\newblock ``Defense against adversarial attacks using feature scattering-based
  adversarial training,''
\newblock in {\em Advances in Neural Information Processing Systems}, 2019.

\bibitem{NEURIPS2019_a2b15837}
Dan Hendrycks, Mantas Mazeika, Saurav Kadavath, and Dawn Song,
\newblock ``Using self-supervised learning can improve model robustness and
  uncertainty,''
\newblock in {\em Advances in Neural Information Processing Systems}, 2019.

\bibitem{abs-2007-02617}
Maksym Andriushchenko and Nicolas Flammarion,
\newblock ``Understanding and improving fast adversarial training,''
\newblock in {\em Advances in Neural Information Processing Systems}, 2020.

\bibitem{abs-2006-07589}
Minseon Kim, Jihoon Tack, and Sung~Ju Hwang,
\newblock ``Adversarial self-supervised contrastive learning,''
\newblock in {\em Advances in Neural Information Processing Systems}, 2020.

\bibitem{8099589}
T.~{Lin}, P.~{Dollár}, R.~{Girshick}, K.~{He}, B.~{Hariharan}, and
  S.~{Belongie},
\newblock ``Feature pyramid networks for object detection,''
\newblock in {\em 2017 IEEE Conference on Computer Vision and Pattern
  Recognition (CVPR)}, 2017.

\bibitem{DBLP:journals/corr/HintonVD15}
Geoffrey~E. Hinton, Oriol Vinyals, and Jeffrey Dean,
\newblock ``Distilling the knowledge in a neural network,''
\newblock {\em CoRR}, vol. abs/1503.02531, 2015.

\bibitem{DBLP:journals/corr/RomeroBKCGB14}
Adriana Romero, Nicolas Ballas, Samira~Ebrahimi Kahou, Antoine Chassang, Carlo
  Gatta, and Yoshua Bengio,
\newblock ``Fitnets: Hints for thin deep nets,''
\newblock in {\em International Conference on Learning Representations}, 2015.

\bibitem{8953432}
T.~{Wang}, L.~{Yuan}, X.~{Zhang}, and J.~{Feng},
\newblock ``Distilling object detectors with fine-grained feature imitation,''
\newblock in {\em 2019 IEEE/CVF Conference on Computer Vision and Pattern
  Recognition (CVPR)}, 2019.

\bibitem{9157636}
K.~{He}, H.~{Fan}, Y.~{Wu}, S.~{Xie}, and R.~{Girshick},
\newblock ``Momentum contrast for unsupervised visual representation
  learning,''
\newblock in {\em 2020 IEEE/CVF Conference on Computer Vision and Pattern
  Recognition (CVPR)}, 2020.

\bibitem{abs-2002-05709}
Ting Chen, Simon Kornblith, Mohammad Norouzi, and Geoffrey~E. Hinton,
\newblock ``A simple framework for contrastive learning of visual
  representations,''
\newblock in {\em Proceedings of the 37th International Conference on Machine
  Learning}, 2020.

\bibitem{abs-2011-10566}
Xinlei Chen and Kaiming He,
\newblock ``Exploring simple siamese representation learning,''
\newblock {\em CoRR}, vol. abs/2011.10566, 2020.

\bibitem{abs-1804-02767}
Joseph Redmon and Ali Farhadi,
\newblock ``Yolov3: An incremental improvement,''
\newblock {\em CoRR}, vol. abs/1804.02767, 2018.

\bibitem{DBLP:journals/corr/SelvarajuDVCPB16}
Ramprasaath~R. Selvaraju, Michael Cogswell, Abhishek Das, Ramakrishna Vedantam,
  Devi Parikh, and Dhruv Batra,
\newblock ``Grad-cam: Visual explanations from deep networks via gradient-based
  localization,''
\newblock {\em Int. J. Comput. Vis.}, 2020.

\end{thebibliography}
\end{document}